\documentclass[conference]{IEEEtran}
\IEEEoverridecommandlockouts
\usepackage{cite}
\usepackage{amsmath,amssymb,amsfonts}
\usepackage{algorithmic}
\usepackage{graphicx}
\usepackage{textcomp}
\usepackage[dvipsnames]{xcolor} 
\def\BibTeX{{\rm B\kern-.05em{\sc i\kern-.025em b}\kern-.08em
    T\kern-.1667em\lower.7ex\hbox{E}\kern-.125emX}}

\usepackage{acronym}
\usepackage{multirow}
\usepackage{hyperref}
\usepackage{enumitem}
\hypersetup{
    colorlinks=False,
    linkcolor=OrangeRed,
    filecolor=OrangeRed,      
    urlcolor=Emerald,
    citecolor=Emerald,
    pdftitle={FedBacys-Extension-Eunjeong},
    }
\usepackage[linesnumbered,lined,boxed,commentsnumbered,ruled,longend]{algorithm2e}
\SetKwIF{If}{ElseIf}{Else}{if}{then}{else if}{else}{}
\SetFuncSty{textsc}
\SetKwInput{KwData}{Input}
\SetKwInput{KwResult}{Output}
\SetKwProg{Init}{Initialize}{}{}
\SetKwFor{ParFor}{for}{do in parallel}{}
\SetKwRepeat{Do}{do}{while}
\SetKw{Return}{return}
\SetKwProg{Fn}{Function}{:}{}
\SetCommentSty{itshape}
\usepackage{bbm}
\usepackage{bbold} 

\newcommand{\cache}[1]{} 

\begin{document}
    \acrodef{fl}[FL]{Federated Learning}
    \acrodef{vaoi}[VAoI]{Version Age of Information}
    \acrodef{ehfl}[EHFL]{Energy-Harvesting Federated Learning}

 \title{Feature-Based Semantics-Aware Scheduling for Energy-Harvesting Federated Learning
\thanks{This work has been supported by the EU through the Horizon Europe/JU SNS project ROBUST-6G (grant no. 101139068).}
}

\author{
    \IEEEauthorblockN{
        Eunjeong Jeong\IEEEauthorrefmark{1},
        Giovanni Perin\IEEEauthorrefmark{2}\thanks{G. Perin is also with the Dept. of Information Engineering (DEI), University of Padova, Padova, Italy.},
        Howard H. Yang\IEEEauthorrefmark{3}, and
        Nikolaos Pappas\IEEEauthorrefmark{1}
    }
    \IEEEauthorblockA{
        \IEEEauthorrefmark{1}\textit{Department of Computer and Information Science}, \textit{Linköping University}, Linköping, Sweden \\
        \IEEEauthorrefmark{2}\textit{Department of Information Engineering}, \textit{University of Brescia}, Brescia, Italy \\
        \IEEEauthorrefmark{3}\textit{ZJU-UIUC Institute}, \textit{Zhejiang University}, Haining, China\\
    Email: \{eunjeong.jeong, nikolaos.pappas\}@liu.se,
    giovanni.perin@unibs.it,
    haoyang@intl.zju.edu.cn
    }
}

\maketitle

\begin{abstract}
    Federated Learning (FL) on resource-constrained edge devices faces a critical challenge: The computational energy required for training Deep Neural Networks (DNNs) often dominates communication costs. However, most existing Energy-Harvesting FL (EHFL) strategies fail to account for this reality, resulting in wasted energy due to redundant local computations. For efficient and proactive resource management, algorithms that predict local update contributions must be devised. We propose a lightweight client scheduling framework using the Version Age of Information (VAoI), a semantics-aware metric that quantifies update timeliness and significance. Crucially, we overcome VAoI's typical prohibitive computational cost, which requires statistical distance over the entire parameter space, by introducing a feature-based proxy. This proxy estimates model redundancy using intermediate-layer extraction from a single forward pass, dramatically reducing computational complexity. Experiments conducted under extreme non-IID data distributions and scarce energy availability demonstrate superior learning performance while achieving energy reduction compared to existing baseline selection policies. Our framework establishes semantics-aware scheduling as a practical and vital solution for EHFL in realistic scenarios where training costs dominate transmission costs.
\end{abstract}

\begin{IEEEkeywords}
Federated learning, semantic communications, energy harvesting, energy efficiency, scheduling algorithms
\end{IEEEkeywords}

\section{Introduction}

\acf{fl}~\cite{konevcny16:FL} has emerged as a powerful framework to enable distributed learning on edge devices while preserving data privacy. However, as \ac{fl} applications increasingly employ Deep Neural Networks (DNNs) over large-scale networks, the dominant constraint has evolved from communication bandwidth to the computational energy required for local training. On resource-constrained edge devices, the energy consumption of backpropagation is substantial, making client scheduling critical for realizing energy-efficient \ac{fl}. A well-designed scheduling policy limits the number of participating clients and prioritizes informative updates, thereby conserving resources while maintaining learning performance.

\ac{ehfl} has been proposed to address energy limitations by leveraging intermittent renewable energy sources. Yet existing \ac{ehfl} literature often assumes that each communication consumes an equal amount of the device's battery and time budget for local training~\cite {chen23:FedSeq, valentedasilva25:FLDA}. This assumption fails in modern edge AI, where a single DNN training epoch consumes significantly more energy than transmitting parameters~\cite{jeong25:fedbacys-extension}. Under this realistic constraint, systems cannot afford to waste energy on training models that yield \emph{redundant} updates. Therefore, blind selection policies based solely on battery levels or random sampling are insufficient; the system must predict a client's contribution \emph{before} initiating expensive training.

This is where semantics-aware metrics offer a promising solution. Unlike traditional metrics focused on throughput or latency, semantics-aware measures evaluate the informational value of updates relative to the learning objective~\cite{kosta17:AoI, OJCOMS-survey, TCOM-Jiping25}. The \ac{vaoi}~\cite{yates21, abolhassani21, buyukates22:vaoi, COMML-Erfan} is particularly promising because it captures both timeliness and relevance of client updates~\cite{delfani25:VAoI}. Prior work has shown that \ac{vaoi}-assisted \ac{fl} can improve robustness and reduce both loss and Age of Information~\cite{hu24:vaoi}. However, existing \ac{vaoi} approaches suffer from a critical limitation: the evaluation of the metric itself is often computationally prohibitive. These methods typically require calculating statistical distances over the entire parameter space, which paradoxically increases the computational burden and renders the approach infeasible for DNNs.

To address this, we propose a lightweight, \ac{vaoi}-based client scheduling framework for \ac{ehfl}. Our approach quantifies update contributions by performing a single forward pass to extract feature vectors from an intermediate layer. This method requires only one forward pass (no backward pass) followed by simple distance calculations, dramatically reducing computational overhead. Experiments demonstrate that \textit{this feature-based \ac{vaoi} approach offers significant advantages under severe data heterogeneity and scarce energy availability}.

The contributions of this study are summarized as follows.
\begin{itemize}
    \item A computation-efficient proxy for model dissimilarity based on intermediate feature extraction, eliminating expensive parameter-space distance calculations.
    \item Demonstration that \ac{vaoi} effectively captures local update significance under realistic constraints where training energy dominates transmission costs.
    \item Experimental validation showing robust learning performance under extreme non-IID data distributions and scarce energy availability, with reduced energy consumption compared to baseline methods.
\end{itemize}

\section{Related Works}
The underlying principle of using lightweight distance metrics to evaluate client importance or dissimilarity is a well-established area of \ac{fl} research. The idea is often framed in the context of personalized Federated Learning (pFL) and Multi-Task Learning (MTL)~\cite{tan2023towards, jeong2022factorized, smith17:FedMTL, marfoq21, huang23, jeong23:pdfl}. For example, in~\cite{jeong2022factorized} the authors use the popular cosine similarity on a factorization of the model parameters of hidden layers, whereas in~\cite{jeong23:pdfl} the Wasserstein distance is computed at the sample level between pairs of clients, but without directly sharing information. While the methodology is similar, we aim to improve the common shared global model instead of creating different versions targeting personalized data distributions.
On similar research lines, distance metrics are also computed based on the dataset labels, whose distribution is assumed to be shared with the server~\cite{fama2024measuring}, loss functions~\cite{tang22:fedcor}, or based on hash sketches mapping the dataset samples into scalars~\cite{liu2023one}. However, we note that the first listed approach partially breaks the privacy-preserving scheme of \ac{fl}, whereas the latter method requires additional computational burden that could be avoided by using the inner feature maps of the local models.

Security mechanisms also use distance and similarity metrics computed on internal representations, model parameters, or loss functions~\cite{blanchard2017machine, fang2020local, sandeepa2024sherpa}. However, these approaches are meant to filter out updates that are considered as outliers, e.g., due to attackers trying to poison the global model. We remark that our goal is fundamentally different, i.e., we want new and different information to join collaborative learning so as to diversify the global model.

Different from what was previously done in the literature, we use the Euclidean distance computed on internal model representations to update the \ac{vaoi}, which is then used to sample clients carrying \emph{useful information} to the global model. Our approach is especially lightweight as it only requires a forward pass of a minibatch of data up to a selected hidden layer. Notably, we show that this procedure yields increased accuracy under non-IID data distributions and/or when clients are constrained by tight energy constraints.


\section{System Model}

\subsection{Federated Learning}
\begin{figure}[t]
    \centering
    \includegraphics[width=\columnwidth]{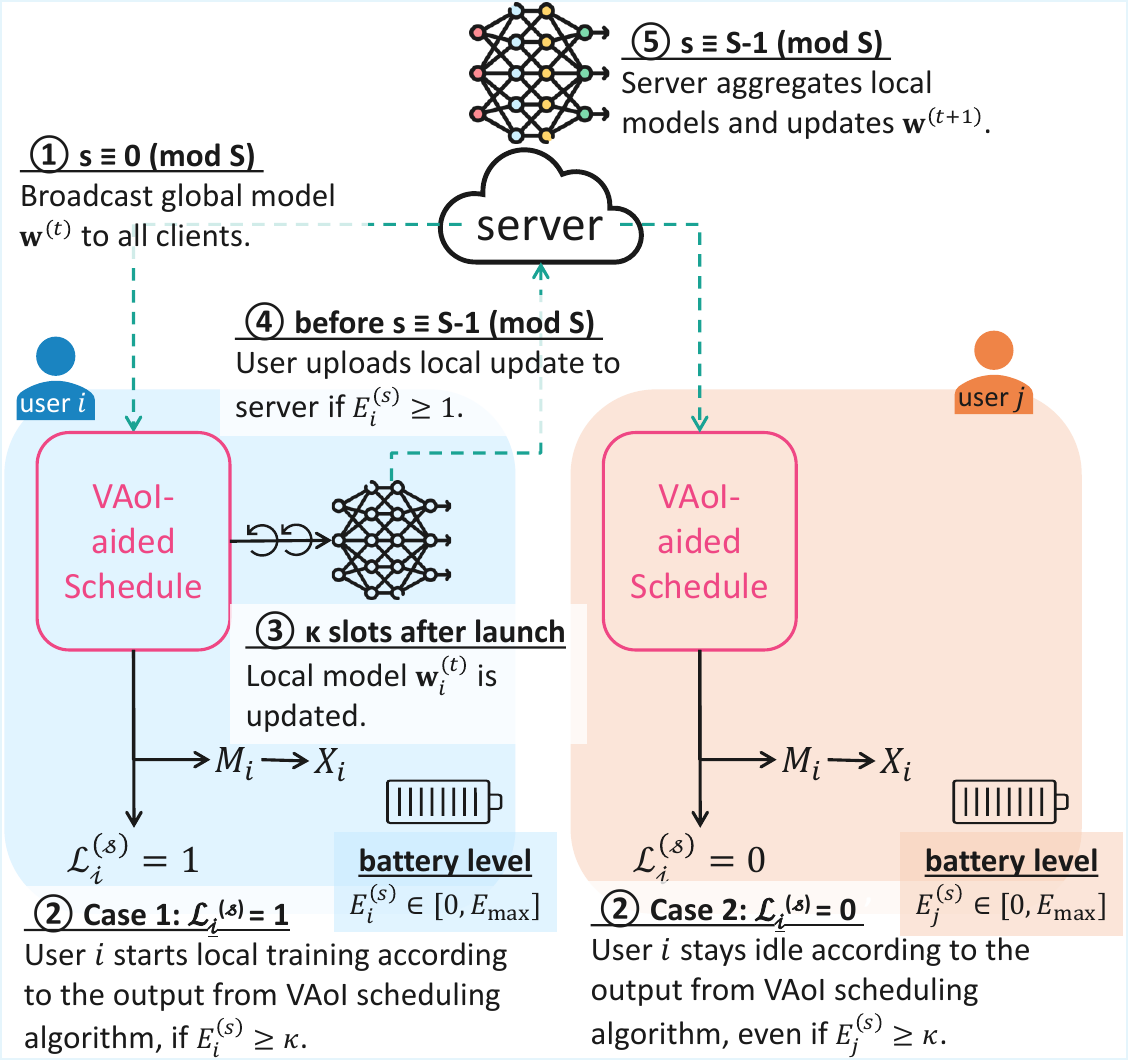}
    \caption{A schematic view of \acl{ehfl}. User~$i$ (left, blue) represents clients participating in local training and model aggregation at epoch~$t$, whereas User~$j$ (right, orange) symbolizes idle clients. Model training and \ac{vaoi}-based participant selection in this figure, depicted as rectangular boxes with magenta edges, is described in Fig.~\ref{fig:VAoI-scheduling}.}
    \label{fig:schematic}
\end{figure}

We consider a distributed network consisting of a central server and $N$ clients, indexed by $i\in[N]=\{1,\ldots,N\}$. The goal of the \ac{fl} process is to find the global model parameters $\mathbf{w}\in\mathbbm{R}^{d_w}$ that minimize the global empirical risk:
\begin{align}\label{eq:objective_function_FL}
    f(\mathbf{w}) = \frac{1}{N}\sum_{i=1}^N f_i(\mathbf{w})\,
\end{align}
where $f_i$ is the local loss function of client $i$, defined over its private and fixed local dataset $\mathcal{D}_i$. We assume a standard Federated Averaging (FedAvg) \cite{mcmahan17:fedavg} as model aggregation method, defined over discrete epochs (global communication rounds) $t\in\{0,\ldots,T-1\}$. An illustration is depicted in Fig.~\ref{fig:schematic}.

\subsection{Version Age of Information}

\acf{vaoi}~\cite{yates21} quantifies the information discrepancy between a source and a receiver, serving as an evaluation metric when the system aims to minimize this information gap. The term ``version'' refers to a discrete counter that tracks the number of updates generated at a node. At $t=0$, the version of the source, $V(t)$, is set to $0$.  When the source's content is updated, its version index increases by 1. Similarly, the $i$-th receiver maintains a local version index, denoted as $V_i(t)$, which updates when its local content is refreshed. The VAoI at node $i$ is $X_i(t) = V(t) - V_i(t)$. This is a general definition independent of the application, and we adapt a modified one tailored to our scenario. 

\ac{vaoi} jointly captures the freshness of the data and the deviation of content between the receiver and the source, which are two semantic properties of information. \ac{vaoi} resets to $0$ when the update is successfully transmitted by the source and incorporated by the receiver, as the entities become synchronized in that instance. Note that it does not increase linearly with time. 

Let $q_i(t)\in\{0,1\}$ refer to a binary indicator with a value $1$ if client~$i$ participates in the global aggregation during round~$t$. At every round~$t$, the \acs{vaoi} of client~$i$ evolves as:
\begin{align}\label{eq:vaoi-default}
    X_i(t+1) = \begin{cases}
        (X_i(t)+1)(1-q_i(t)), & d(\mathbf{w}^{(t)}, \mathbf{w}_i^{(t)})\geq\mu\\
        X_i(t)(1-q_i(t)), & \text{otherwise}
    \end{cases}
\end{align}
where $\mu$ is the predetermined threshold. The measure $d(\mathbf{w}^{(t)}, \mathbf{w}_i^{(t)})$ in (\ref{eq:vaoi-default}) is the statistic distance (dissimilarity) between the global model $\mathbf{w}^{(t)}$ and the current local model $\mathbf{w}_i^{(t)}$, capturing the model disparity. 

\subsection{Energy Harvesting}

\begin{figure}[t]
    \centering
    \includegraphics[width=\columnwidth]{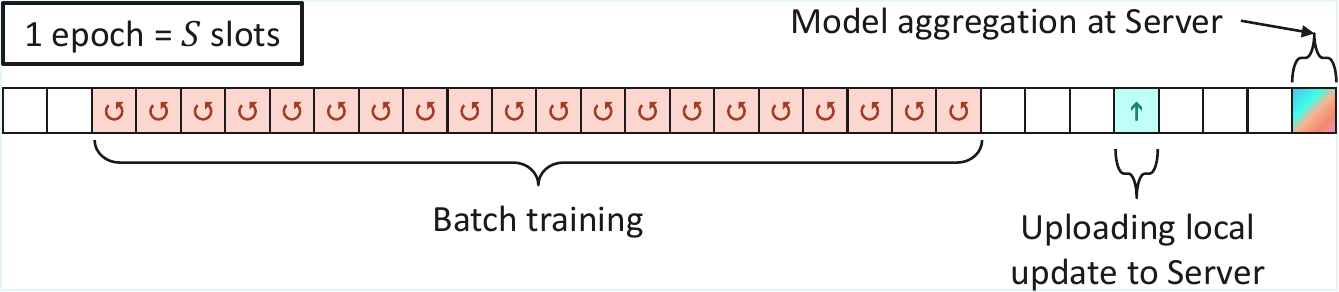}
    \caption{Relation between epochs and slots with an exemplary case of a client allocating slots for scheduled and/or available actions. $S=30$ and $\kappa=20$ are assumed in this example.}
    \label{fig:epoch_and_slot}
\end{figure}
\begin{figure}[t]
    \centering
    \includegraphics[width=0.75\columnwidth]{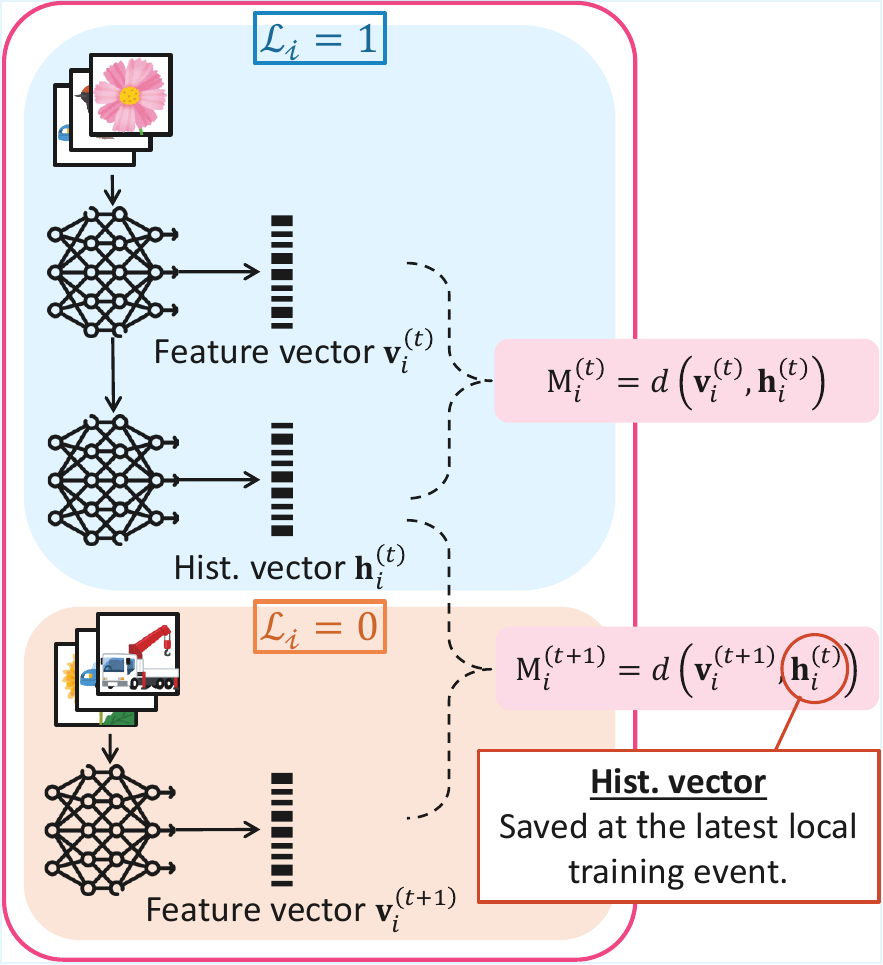}
    \caption{VAoI-aided participant selection. The figure illustrates an exemplary case in which a user is selected for local training at epoch~$t$ (upper block, blue) and keeps the local model untrained at epoch~$t+1$ (lower block, orange). If the most recent local update was done $\tau_i$ epochs ago, $M_i^{(t)}$ is obtained by measuring distance between $\mathbf{v}_i^{(t)}$ and $\mathbf{h}_i^{(t-\tau_i)}$.}
    \label{fig:VAoI-scheduling}
\end{figure}

We consider a system operating over discrete time slots $s\in\{0,\ldots,ST-1\}$, where $S$ is the number of slots per epoch and $T$ is the number of epochs, as illustrated in Fig.~\ref{fig:epoch_and_slot}. Each user $i$ relies on a rechargeable battery with current energy level $E_i^{(s)}$. The battery levels depend on the stochastic energy harvesting and the users' operational decisions.

At the beginning of each time slot $s$, user $i$ may harvest a unit of energy from the environment. We denote this event by the binary variable $\mathcal{C}_i^{(s)}$, where $\mathcal{C}_i^{(s)}=1$ implies the battery recharges by one unit, and $\mathcal{C}_i^{(s)}=0$ implies no energy is harvested. The probability of harvesting is given by $p_\text{bc} = \text{Pr}[\mathcal{C}_i^{(s)}=1]$, thus $\mathcal{C}_i^{(s)}$ following a Bernoulli distribution.

If the user is not currently locked into a long-term task (i.e., they are available),
they choose among three actions: (1) staying idle (1 slot, 0 energy), (2) transmission (1 slot, 1 energy), or (3) local training ($\kappa$ slots, $\kappa$ energy). We define binary decision variables $\mathcal{T}_i^{(s)}$ and $\mathcal{L}_i^{(s)}$  for transmission and training engagement, respectively. Since a user cannot simultaneously transmit and initiate training, the decision variables satisfy the constraint $\mathcal{T}_i^{(s)} + \mathcal{L}_i^{(s)} \leq 1$.


The battery level evolves according to the chosen action and the harvested energy. For single-slot actions (idle or transmission), the battery level at the next step is:
\begin{equation}
    E_i^{(s+1)} = \max(E_i^{(s)}-\mathcal{T}_i^{(s)},0)+\mathcal{C}_i^{(s)},
\end{equation}

On the other hand, in the case of multi-slot action as local training, the user remains busy for $\kappa$ consecutive slots if it initiates training at slot $s$. The battery level is updated only after the task completes at slot $s+\kappa$:
\begin{equation}
    E_i^{(s+\kappa)} = \max(E_i^{(s)}-\kappa,0)+\sum_{s'=s}^{s+\kappa-1}\mathcal{C}_i^{(s')}.
\end{equation}
Note that if the user remains idle ($\mathcal{T}_i^{(s)} = \mathcal{L}_i^{(s)} = 0$), the battery simply accumulates harvested energy.

The system enforces strict energy causality. A user can undertake an action only if their current battery level $E_i^{(s)}$ is sufficient to fully cover the energy cost of that action, otherwise the action is denied. Thus, transmission is only possible if $E_i^{(s)} \geq 1$, and local training is only possible if $E_i^{(s)} \geq \kappa$. 


\section{Version Age of Information in Energy-Harvesting Federated Learning}

\begin{algorithm}[t]
    \SetAlgoLined \SetAlgoNoEnd
    \caption{\label{alg:FEVA} \acs{ehfl} with lightweight \acs{vaoi} for client selection.}
    \DontPrintSemicolon
    \KwData{Total epochs $T$, number of slots per epoch $S$, user set $[N]=\{1,\cdots,N\}$, battery charging probability $p_\text{bc}$, learning rate $\gamma$,
    initial battery level $E_i^{(0)}=0$, local training model $\mathbf{w}_i^{(t)}=\mathbb{0}$, local dataset $\{\mathcal{D}_i\}$ $\ \forall i\in[N]$}
    \KwResult{$\{\mathbf{x}_t : \forall t\}$}   

    \For{$s=0,\cdots,ST-1$}{
        $t\leftarrow \lfloor s/S\rfloor$ \tcp*{index of current epoch}
        \ParFor{each $i\in [N]$}{
            \If{$\mathcal{C}_i^{(s)}=1$}{
                $E_i^{(s)}\leftarrow E_i^{(s)}+1$ \tcp*{Charge battery}
            }
            \If{$\mathcal{L}_i^{(s)}=1$}{
                $\mathbf{w}_i^{(t,b+1)} \leftarrow$ \textsc{BatchTrain}($\mathbf{w}_i^{(t,b)}, \mathcal{D}_i, b$)\;
            }
            \ElseIf{$i$ has an unsent $\mathbf{w}_i$ and $E_i^{(s)}\geq 1$}{
                $i$ uploads $\mathbf{w}_i$ (message) to the server.\;
            }
        }
        \If{$s\equiv 0\ (\text{mod }S)$}{
            \For{each $i\in [N]$}{
                $\mathbf{w}_i^{(t)} \leftarrow \mathbf{w}^{(t)}$ \tcp*{Server broadcasts global model.}   
            }
            $\mathcal{N}(t),\{X_i(t+1)\} \leftarrow $ \textsc{ClientSelect}($\{X_i(t)\},\{\mathbf{w}^{(t)}\},k$) \tcp*{Alg. \ref{alg:VAoI-Schedule}} 
        }
        \ParFor{each $i\in \mathcal{N}(t)$}{
            \If{$i$ meets conditions to start training}{
                Set $\mathcal{L}_i^{(s')}=1$ for $s'\in\{s,\ldots,s+\kappa-1\}$\;
                $\mathbf{w}_i^{(t,0)}\leftarrow \mathbf{w}_i^{(t)}$\;
               $\mathbf{w}_i^{(t,1)}\leftarrow$\textsc{BatchTrain}($\mathbf{w}_i^{(t,0)}, \mathcal{D}_i, 0$)\;
            }
        }
        \If{$s\equiv S-1\ (\text{mod }S)$}{
            $\beta_i^{(t)}\leftarrow\frac{|\mathcal{D}_i|}{\sum_{i=1}^N |\mathcal{D}_i|}$ \tcp*{superposition weight}
            $\mathbf{w}^{(t+1)}\leftarrow \sum_{i\in \mathcal{N}(t)}\beta_i^{(t)}\mathbf{w}_i^{(t)}$ \tcp*{Aggregation}
        } 
        }
    \Return $\mathbf{w}^{(T)}$\;
    \vspace{0.3em}
    \SetKwFunction{BTrn}{BatchTrain}
    \Fn{\BTrn{$\mathbf{w}_i^{(t)}, \mathcal{D}_i, b$}}{
         $\mathcal{B}_i^{(t,b)}\leftarrow$ \{Minibatch sampled from $\mathcal{D}_i$\} \;
         $\mathbf{w}_i^{(t,b+1)} \leftarrow \mathbf{w}_i^{(t,b)} - \gamma\nabla f_i(\mathbf{w}_i^{(t,b)}; \mathcal{B}_i^{(t,b)})$ \;
         \If{$b=\kappa-1$}{
            $i$ locally saves $\mathbf{w}_i^{(t,\kappa)}$ as message.\;
            $i$ computes and locally saves $\mathbf{h}_{i}^{(t)}$  as in (\ref{eq:hist_moment}). \;
         }   
    \Return{$\mathbf{w}_i^{(t,b+1)}$}}
\end{algorithm} 

Encoding feature vectors for evaluating model dissimilarity takes the concept of checking feature map difference and makes it statistically robust and hyper-lightweight. Instead of calculating the full L2 distance for all feature vectors, the system focuses on the statistical properties of features generated by the local and global models.

The model dissimilarity metric is used as follows.
\begin{enumerate}[label=\arabic*), leftmargin=*]
    \item \textbf{Client's inference:} Client~$i$ runs a single forward pass on a small local batch $\mathcal{B}_i$ using the global model $\mathbf{w}^{(t)}$.
    \item \textbf{Feature vector extraction:} Client~$i$ extracts the feature vector from a chosen hidden layer $\mathbf{z}$.
    \item \textbf{Statistical distance check:} 
    Client~$i$ compares the global model-generated feature vectors with the historical moments it has recorded.
    \begin{equation}\label{eq:stat_distance}
        M_i^{(t)} = \Big\|\frac{1}{|\mathcal{B}_i|}\big(\mathbf{z}(\mathbf{w}^{(t)};\mathcal{B}_i)\big) - \mathbf{h}_{i}^{(t-\tau_i)} \Big\|_2
    \end{equation}
    where $\mathbf{h}_{i}^{(\cdot)}$ is the average feature vector (mean of the hidden layer output) stored by client $i$, computed at the latest time slot when client~$i$ was trained (i.e., when $\mathbf{w}_{i}$ was last created), denoted by $t-\tau_i$.
    When $\mathcal{L}_i^{(s')}=1$ for slots ranging $s'\in\{s,\ldots,s+\kappa-1\}$ that lays on epoch $t$, the historical moment vector of client $i$ is obtained by feeding the training batches $\mathcal{B}_i^{(t,s')}$:
    \begin{equation}\label{eq:hist_moment}
        \mathbf{h}_i^{(t)} = \frac{1}{|\mathcal{D}_i|} \sum_{s'=s}^{s+\kappa-1} \mathbf{z}(\mathbf{w}_i^{(t,s')}; \mathcal{B}_i^{(t,s')}),
    \end{equation}
    where $|\mathcal{D}_i|$ is the number of training samples in $\mathcal{D}_i$.
    \item \textbf{Staleness evaluation (\ac{vaoi}):} The \ac{vaoi} ($X_i(t)$) is updated using the standard evolution rule in Eq.~(\ref{eq:vaoi-default}), with the following substitution: The expensive full model dissimilarity, $d(\mathbf{w}^{(t)}, \mathbf{w}_i^{(t)})$, is replaced by the lightweight feature-based distance, $M_i^{(t)}$, calculated in Step 3.
    \begin{align}\label{eq:vaoi}
        X_i(t+1) = \begin{cases}
            (X_i(t)+1)(1-q_i(t)), & M_i^{(t)}\geq\mu\\
            X_i(t)(1-q_i(t)), & \text{otherwise}.
        \end{cases}
    \end{align}
    If Client $i$ is selected to participate ($q_i(t)=1$), the age resets to zero, as the client synchronizes with the new global model. Otherwise, the age only increments if the computed distance $M_i^{(t)}$ exceeds the predetermined threshold $\mu$, indicating that the current global update is semantically significant to client $i$.
\end{enumerate}

The comprehensive execution flow of the proposed scheme is outlined in Alg.~\ref{alg:FEVA}. Each client~$i$ begins batch training over its local model only if it satisfies the following conditions, which are assessed in line 15 of Alg.~\ref{alg:FEVA}:
\begin{itemize}
    \item Priority based on version age: $i\in\mathcal{N}(t)$, where the participation group $\mathcal{N}(t)$ is formed according to the function \textsc{ClientSelect} in Alg.~\ref{alg:VAoI-Schedule}.
    \item No pending update: the user does not yet have a local update $\mathbf{w}_i$ to transmit to the server.
    \item Sufficient battery: $E_i^{(s)}\geq \kappa$.
\end{itemize}

\begin{algorithm}[t]
    \SetAlgoLined \SetAlgoNoEnd
    \caption{\label{alg:VAoI-Schedule} \ac{vaoi}-based client selection.}
    \DontPrintSemicolon
    \KwData{global model $\mathbf{w}^{(t)}$, Upper bound for choosing participants with large version age $k$, Version age at the current epoch $X_i(t)$ $\ \forall i\in[N]$}
    
    \SetKwFunction{Scheduling}{ClientSelect}
    \SetKwProg{Fn}{Function}{:}{end}

    \Fn{\Scheduling{$\{X_i(t)\}, \{\mathbf{w}^{(t)}\}, k$}}{
        Compute probabilities $\{p_i\}\leftarrow \frac{X_i(t)}{\sum_{j=1}^N X_j(t)}$. \;
        $\mathcal{N}(t) \leftarrow$ \{Subset of $k$ clients with largest $p_i$.\}\;
        \For{each $i\in[N]$}{
            \If{$i\in \mathcal{N}(t)$}{$q_i(t) \leftarrow 1$}
        $\mathbf{v}_i^{(t)} \leftarrow \frac{1}{|\mathcal{B}_i|} \mathbf{z}(\mathbf{w}^{(t)};\mathcal{B}_i)$ \tcp*{Forward pass; Return avg. feature vector}
        $M_i \leftarrow \|\mathbf{v}_i^{(t)}-\mathbf{h}_{i}^{(t-\tau_i)}\|_2$ \;
        Find $X_i(t+1)$ from Eq.~(\ref{eq:vaoi}) \;
        }
    }\Return $\mathcal{N}(t), \{X_i(t+1)\}$\; 
\end{algorithm} 


\section{Numerical Results}

\begin{figure*}[t]
    \centerline{
    \includegraphics[width=\linewidth]{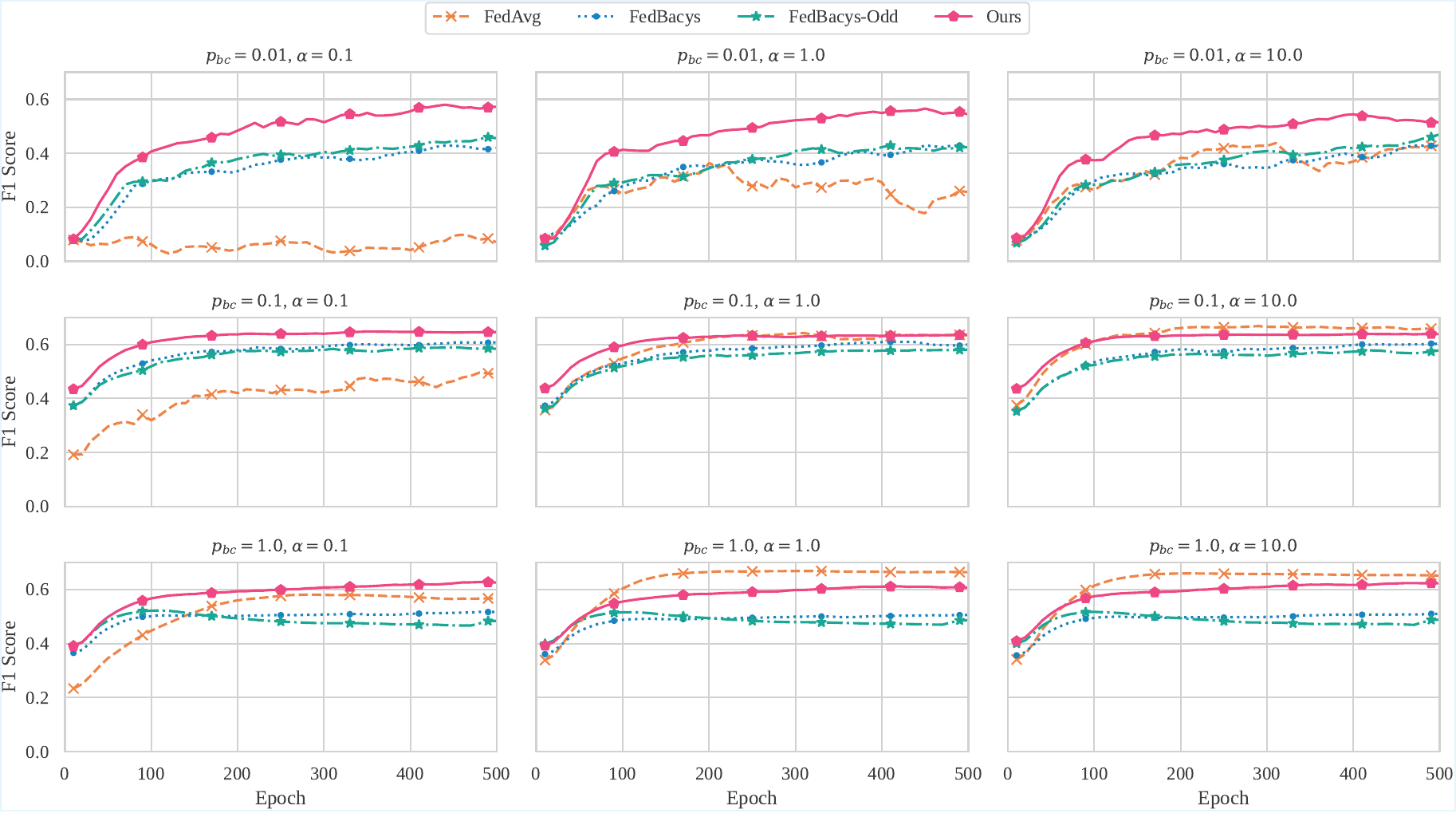}}
    \caption{F1 score with respect to number of epochs. }
    \label{plot:f1}
\end{figure*}

\begin{figure*}[t]
    \centering
    \includegraphics[width=0.95\linewidth]{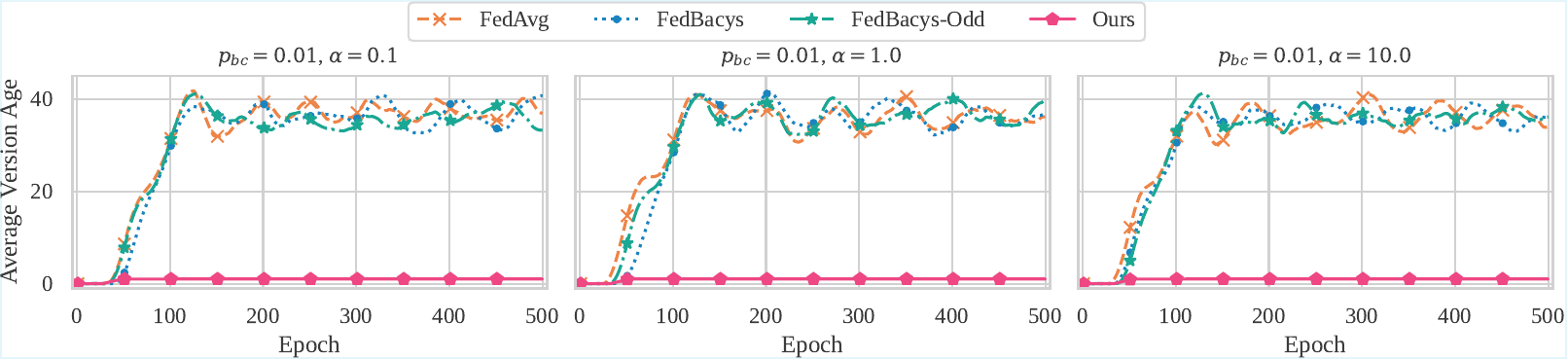}
    \caption{Average version age across all clients with respect to epochs.}
    \label{fig:average_version_age}
\end{figure*}

We evaluate the proposed scheme on an image classification task using the CIFAR-10 dataset. Each model is a convolutional neural network consisting of six convolutional layers, three max-pooling layers, and three fully-connected layers. We consider $N=100$ clients, each of which possesses 300 training samples. To simulate non-IID data distribution, we partition the dataset following a Dirichlet distribution with concentration coefficient $\alpha=\{0.1, 1.0, 10.0\}$, where smaller $\alpha$ indicates more severe heterogeneity. stochastic gradient descent with a learning rate $\gamma=0.01$ and cross-entropy loss over $T=500$ global rounds, each comprising $S=30$ time slots.
We extract representations from the output layer of the local model for the lightweight \ac{vaoi} calculation. This layer generates a feature vector with 10 elements, corresponding to the number of classes in CIFAR-10.
For energy harvesting, we vary the battery charging probability $p_\text{bc} \in \{0.01, 0.1, 1.0\}$. We set the computational cost of a single local training to $\kappa=20$ battery units, while uplink transmission consumes 1 battery unit. The maximum battery capacity is $E_{\max}=\kappa+5=25$ units, and initial battery is set $E_i^{(0)}=0$ for all $i\in[N]$.

We implement the following algorithms as benchmarks:
\begin{LaTeXdescription}
    \item[FedAvg~\cite{mcmahan17:fedavg}:] A greedily energy-aware baseline. Battery evolution follows a Bernoulli process. Clients operate using a greedy strategy, launching local training as soon as the minimum energy requirement ($\kappa$ units) is met, without consideration of future energy needs or data quality. 
    \item[FedBacys~\cite{jeong25:FedBacys}:] A procrastination-based, deadline-driven scheme. Clients are organized into cyclical groups, and each group must meet a strict transmission deadline. Clients procrastinate initiating local training until the last possible time slot to maximize harvested energy. 
    \item[FedBacys-Odd~\cite{jeong25:fedbacys-extension}:] An energy-efficient variation of FedBacys designed for scenarios where training energy dominates communication. Clients utilize the same launching criteria as FedBacys: (i) sufficient battery, (ii) no pending updates, and (iii) deadline observance. Additionally, clients launch training only if an internal counter, tracking opportunities met by criteria (i)-(iii), is odd-numbered. The odd-chance rule skips every other valid training opportunity to save energy.
\end{LaTeXdescription}

We evaluate the performance of the considered schemes based on two key metrics: (1) Learning performance measured by the F1 score on the global test set, and (2) energy efficiency measured by the total cumulative battery units consumed by the network throughout the training process.

Fig.~\ref{plot:f1} demonstrates the advantages of leveraging \ac{vaoi} under severe heterogeneity (small $\alpha$, top row) and scarce energy (small $p_{bc}$, left column) conditions. Our scheme balances energy efficiency with learning performance by prioritizing semantically informative clients, ensuring that critical updates are not missed when energy is limited.

The average \ac{vaoi} for all clients in each global round is shown in Fig.~\ref{fig:average_version_age}, which shows that our proposed scheme maintains the lowest average version age among all comparisons. This result is consistent with its design objective, as the selection policy explicitly prioritizes clients with higher version ages (staleness) and high significance to global update. This emphasizes that \ac{vaoi} is an effective metric for guiding FL networks toward a robust and sustainable trajectory. 

\begin{figure}[t] 
    \centering
    \includegraphics[width=\columnwidth]{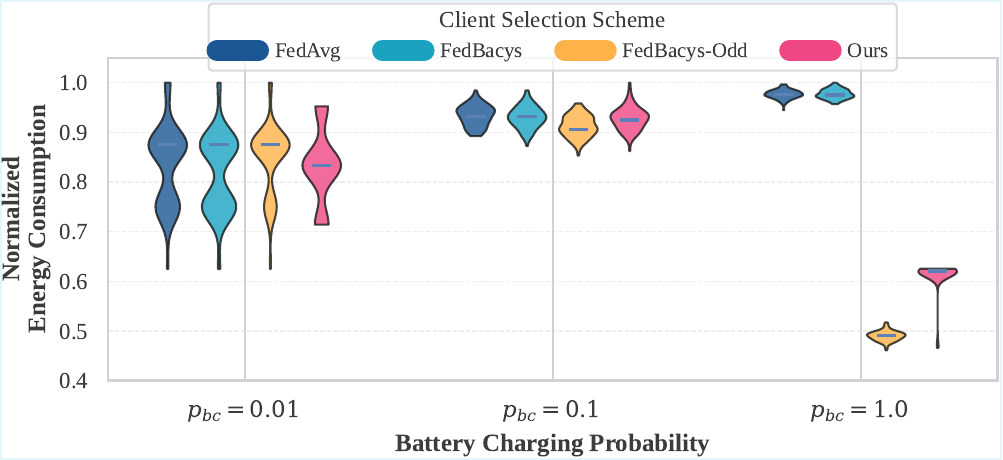}
    \caption{Network-wide energy consumption ($\alpha=0.1, k=10, \mu=0.5$). Normalization is done within each $p_{bc}$ group by dividing all values by the max value across all 4 schemes in each group.} 
    \label{fig:energy_consumption}
\end{figure}

Fig.~\ref{fig:energy_consumption} illustrates the total energy consumption across different schemes. We observe that the degree of data heterogeneity ($\alpha$) does not impact the energy consumption patterns of any scheme, as energy expenditure is primarily driven by participation frequency rather than data distribution. Thus, a notable performance gap appears when the battery charging probability ($p_{bc}$) is high. Our proposed VAoI-aided scheme reduces energy consumption by up to 37\% compared to the baseline FedAvg. While the purely battery-aware strategy (FedBacys-Odd) achieves slightly lower total energy consumption than our algorithm, it proves vulnerable under high data heterogeneity since it lacks a mechanism to account for the utility of local updates.

\section{Conclusions}
We addressed a critical challenge in \acf{ehfl} concerning the dominance of local training costs over communication costs in Deep Neural Network scenarios. We proposed a lightweight, \ac{vaoi}-driven scheduling framework that utilizes a feature-based proxy to estimate the semantic utility of updates via a single forward pass. By effectively filtering out redundant computations, our approach achieves significant energy savings and maintains robust convergence under extreme non-IID and scarce energy conditions, outperforming standard baselines, and establishing semantics-aware selection as a vital mechanism for sustainable \ac{fl} deployment.



\bibliographystyle{ieeetr}  
\bibliography{IEEEabrv, DAabrv, ref}

\end{document}